# A Hybrid Meta-Learning and Multi-Armed Bandit Approach for Context-Specific Multi-Objective Recommendation Optimization

Tiago Cunha
tsacunha@expediagroup.com
Expedia Group

Andrea Marchini
amarchini@expediagroup.com
Expedia Group

## ABSTRACT

Recommender systems in online marketplaces face the challenge of balancing multiple objectives to satisfy various stakeholders, including customers, providers, and the platform itself. This paper introduces Juggler-MAB, a hybrid approach that combines meta-learning with Multi-Armed Bandits (MAB) to address the limitations of existing multi-stakeholder recommendation systems. Our method extends the Juggler framework, which uses meta-learning to predict optimal weights for utility and compensation adjustments, by incorporating a MAB component for real-time, context-specific refinements. We present a two-stage approach where Juggler provides initial weight predictions, followed by MAB-based adjustments that adapt to rapid changes in user behavior and market conditions. Our system leverages contextual features such as device type and brand to make fine-grained weight adjustments based on specific segments. To evaluate our approach, we developed a simulation framework using a dataset of 0.6 million searches from Expedia's lodging booking platform. Results show that Juggler-MAB outperforms the original Juggler model across all metrics, with NDCG improvements of 2.9%, a 13.7% reduction in regret, and a 9.8% improvement in best arm selection rate.

## 1 INTRODUCTION

Traditional recommender systems often focus solely on user satisfaction. However, in many real-world applications, particularly in online marketplaces, multiple stakeholders' interests need to be considered [1, 2]. These stakeholders typically include users (customers), item providers (e.g., hotel owners), and the platform itself. Multi-stakeholder recommender systems aim to balance these diverse and sometimes conflicting objectives [3, 4].

The Juggler framework [5] was introduced to address this multi-stakeholder recommendation problem by using meta-learning [6, 7] to predict optimal weights for utility and compensation adjustments in real-time scoring. Deployed in production, Juggler has been an integral part of the Lodging Ranking stack at Expedia. However, Juggler's reliance on a pre-configured set of five options for relevance and compensation limits its ability to fine-tune recommendations for specific contexts. Additionally, its infrequent training cycles make it less responsive to rapid changes in traffic patterns across different segments.

To address these limitations, we propose a two-step approach that combines meta-learning (Juggler) with Multi-Armed Bandits for multi-stakeholder recommendations for real-time weight adjustments. This approach aims to: 1) provide more granular weight adjustments based on specific segments (e.g., device type, brand) and 2) adapt quickly to changes in traffic patterns without requiring frequent retraining of the main Juggler model.

Our research questions are:

- Can the integration of MAB with Juggler improve the performance and adaptability of multi-stakeholder recommendations in online marketplaces?
- Are contextual features useful to improve the MAB's effectiveness at making the right decisions?

The rest of this paper is organized as follows: Section 2 presents the related work, while Section 3 introduces the proposed hybrid solution. Section 4 covers the experimental setup used to validate the proposal and while Section 5 reports the results to the research questions. Lastly, Section 6 highlights the main conclusions and avenues for future work.

## 2 BACKGROUND AND RELATED WORK

The Juggler framework [5] was introduced as a meta-learning approach to address the multi-stakeholder recommendation problem. It dynamically predicts the ideal weights for utility (user relevance) and compensation (platform revenue) for each search query. The meta-model leverages a collection of historical search queries and learns the mapping between the search context and the ideal utility and compensation weights, learned via offline simulations. Juggler selects from five pre-configured options, each representing a different balance between relevance and compensation: 1) Lower relevance, lower compensation, 2) Lower relevance, higher compensation, 3) Neutral relevance, neutral compensation, 4) Higher relevance, lower compensation and 5) Higher relevance, higher compensation. The pre-configured options refer to sections of the search space which are explored to identify different directions of improvement, while reducing the number of options to ultimately choose from. It is noteworthy that although the pre-configured options are fixed, the actual instantiation of weights for each option depends on the ranking problem characteristics and Juggler framework hyper-parameters. While Juggler has shown success in production, its reliance on these fixed options and infrequent training cycles limits its adaptability to rapid changes in user behavior and market conditions.

Multi-Armed Bandits (MAB) are a class of reinforcement learning algorithms that balance exploration and exploitation in decision-making processes [8]. In the context of recommender systems, MABs have been used to address the exploration-exploitation dilemma and to adapt to changing user preferences [9, 10].

The integration of meta-learning and bandit algorithms has been explored in other domains, such as algorithm selection [11] and hyperparameter optimization [12]. Our work extends these ideas to the realm of multi-stakeholder recommender systems, addressing the unique challenges posed by the dynamic nature of online marketplaces.

Several studies have addressed the challenge of balancing multiple objectives in recommender systems. Rodriguez et al. [13] proposed a multi-objective optimization approach for job recommendations. Nguyen et al. [14] introduced a multi-objective learning to re-rank approach to optimize online marketplaces for multiple stakeholders. Sürer et al. [15] explored multi-stakeholder recommendation with provider constraints. These approaches provide valuable insights into balancing multiple objectives, but our proposed method aims to extend their capabilities by combining meta-learning with multi-armed bandits for enhanced adaptability in dynamic online marketplaces.

Recent developments in industry have led to the creation of self-service platforms for deploying contextual bandits, such as AdaptEx [16]. These platforms provide powerful tools for optimizing user experiences at scale, which we leverage in our hybrid approach to combine the strengths of meta-learning and MAB algorithms. To evaluate our approach, we utilized a custom simulation framework based on real-world data from an online travel marketplace. This allowed us to assess the performance of our system in a controlled yet realistic setting, similar to other sophisticated simulation environments used for complex recommender systems [17].

## 3 JUGGLER WITH MAB

We present a hybrid approach that combines the Juggler framework's meta-learning capabilities [5] with a MAB system powered by the AdaptEx SDK [16]. This approach, which we call "Juggler-MAB" aims to address the limitations of the original Juggler system while leveraging the adaptive capabilities of contextual bandits. The Juggler-MAB system operates in two stages:

(1) **Juggler Stage**: The meta-learning model predicts initial utility and compensation weights based on search context.
(2) **MAB Stage**: A contextual MAB refines these weights in real-time based on user interactions and contextual features, if applicable.

The Juggler framework selects from five pre-configured options for utility and compensation weights, providing a coarse adjustment of the recommendation strategy based on the search context. These options range from lower relevance and compensation to higher relevance and compensation, as described in [5] and aim to tackle the main issues in multi-objective optimization.

The MAB component introduces fine-grained adjustments to the Juggler-predicted weights. Each arm of the bandit represents a small corrective measure to be applied to the utility and compensation weights to improve relevance.

The key features of our MAB implementation include:

(1) **Contextual arms**: The contextual bandits consider contextual features (e.g., device type, brand) when selecting arms.
(2) **Reward function**: We use Normalized Discounted Cumulative Gain (NDCG) as a proxy for Conversion Rate, allowing for offline simulation and evaluation.
(3) **Exploration strategy**: We employ epsilon-greedy and Thompson Sampling for its ability to balance exploration and exploitation effectively [8].

The integration of Juggler and MAB is achieved through an additive approach in the scoring function:

$$sortScore = (w_{utility}^{Juggler} + w_{utility}^{MAB}) \cdot utilityScore \\ + (w_{comp}^{Juggler} + w_{comp}^{MAB}) \cdot compensationScore \quad (1)$$

where $w_{utility}^{Juggler}$ and $w_{comp}^{Juggler}$ are the weights predicted by Juggler, and $w_{utility}^{MAB}$ and $w_{comp}^{MAB}$ are the corrective weights determined by the MAB.

We formulate our contextual MAB problem as follows: let $\mathcal{A}$ be the set of arms, where each arm $a \in \mathcal{A}$ represents a pair of corrective weights ($w_{utility}^{MAB}, w_{compensation}^{MAB}$). The context $x_t \in \mathcal{X}$ at time $t$ includes features such as device type or brand. The reward $r_t$ is defined as the NDCG of the resulting ranking. The goal is to find a policy $\pi : \mathcal{X} \to \mathcal{A}$ that maximizes the expected cumulative reward:

$$\max_{\pi} \mathbb{E}\left[\sum_{t=1}^{T} r_t(x_t, \pi(x_t))\right] \quad (2)$$

where $T$ is the time horizon.

We explored various methods to combine Juggler's predictions with MAB corrections, ultimately settling on the additive approach described above. We carefully selected contextual variables that would help identify under-performing segments in the Juggler model, such as device type and brand. Balancing multiple objectives in a single reward function required careful consideration. We chose NDCG as an initial approach, with plans to explore more complex multi-objective reward functions in future work.

To evaluate our hybrid approach, we developed a custom simulation tool that allows us to test various configurations offline using historical data. This simulator, built on top of real Expedia data, enables us to:

(1) Replay historical searches and user interactions
(2) Apply the Juggler-MAB model to generate new rankings
(3) Evaluate the performance using both immediate (e.g., clicks) and delayed (e.g., bookings) feedback

The simulation framework provides a safe environment to test and refine our approach before considering online deployment.

## 4 EXPERIMENTAL SETUP

We used a dataset of 0.6 million searches from Expedia's lodging booking platform, covering a period of 31 consecutive days. The dataset includes features such as device type, brand, destination, and historical user interactions (i.e. clicks and bookings).

We implemented our simulation framework using the lodging MAB simulator tool, which allows us to replay historical data and evaluate different bandit configurations. The simulator was run on AWS i3.4xlarge EC2 instances.

We compared several variants of the proposed Juggler-MAB hybrid approach against the original Juggler model [5]. We tested several MAB algorithms, ranging from classical (i.e. no contextual features) to contextual bandits:

- Gaussian Thompson (GT): a classical bandit using Thompson Sampling assuming a Gaussian Distribution of reward value.
- $\epsilon$-greedy: a classical bandit using a vanilla implementation of the canonical algorithm. We have used $\epsilon = 0.1$ and $\epsilon = 0.3$.
- Recursive Least Squares with Thompson Sampling (RLS): a contextual bandit using a linear model with a vector of means and a matrix of variances-covariances.

Given that the Juggler model already is optimized towards the multi-objective domain, we rely simply on the relevance metric Normalized Discounted Cumulative Gain (NDCG) to determine how well can MAB algorithms correct towards relevance and expected conversion rate improvement. The experiments use the actual production Juggler model predictions for each search used in the experiments. This improves the reliability of Juggler's predictions, which in turn leads to more robust estimates of the MAB's effect. We then implemented the MAB component using the AdaptEx SDK [16], with the following configuration:

- Arm space: we explore 3 different values for each arm, respectively $w_{utility}^{MAB} \in \{-0.3, 0.0, 0.3\}$ and $w_{comp}^{MAB} \in \{-0.2, 0.0, 0.2\}$. The selected weights ensure non-zero weights for both scores.
- Contextual features: several low cardinality categorical search features were tested, with 3 being identified as the most important: brand, user device and geographical categorization of the search destination, e.g. neighborhood vs city.
- Exploration strategy: Thompson Sampling
- Reward: Normalized Discounted Cumulative Gain (NDCG)

We ran simulations for 31 days, processing on average 22 thousand searches per day. For each search, we recorded the selected arm, the resulting ranking, and the user feedback and the contextual features, if applicable. We computed the reward on a daily basis and used them to update the bandit for the following day.

## 5 RESULTS AND DISCUSSION

Table 1 summarizes the results for all bandits and the Juggler model baseline. For each bandit, we report the average reward, regret and the percentage of best arm selections across all searches. The best results per metric are highlighted in bold. Notice that regret is best when lowest, the remaining metrics are better when maximized.

Our Juggler-MAB hybrid approach outperformed the Juggler baseline across all metrics for all bandits proposed. The NDCG improvements range from +0.8% for GT bandit, all the way to +2.9% in several RLS bandits. In terms of regret, we achieve a reduction of 13.7% and an improvement in best arm selection rate of 9.8%.

| Bandit | avg(reward) | avg(regret) | best arm % |
|---|---|---|---|
| Juggler | 0.1776 | 0.0373 | 0.7515 |
| GT | 0.1791 | 0.0358 | 0.7866 |
| $\epsilon$-greedy (0.3) | 0.1811 | 0.0339 | 0.8095 |
| $\epsilon$-greedy (0.1) | 0.1824 | 0.0325 | 0.8218 |
| $RLS_{brand}$ | **0.1827** | **0.0322** | **0.8252** |
| $RLS_{device}$ | 0.1822 | 0.0327 | 0.8200 |
| $RLS_{geo}$ | 0.1825 | 0.0325 | 0.8228 |
| $RLS_{geo, brand}$ | **0.1827** | 0.0323 | 0.8246 |
| $RLS_{device, brand}$ | **0.1827** | **0.0322** | 0.8228 |
| $RLS_{geo, device}$ | **0.1827** | **0.0322** | 0.8247 |
| $RLS_{geo, device, brand}$ | 0.1826 | 0.0323 | 0.8246 |

Table 1: Aggregated reward, regret and the percentage of best arm selections results for all bandits and baselines.

The $\epsilon$-greedy algorithms provide very strong baselines, especially when $\epsilon = 0.1$. GT bandit is clearly the worst bandit, but yet useful since it outperforms the baseline. Among the contextual bandits, the best one across all metrics is the $RLS_{brand}$. Interestingly, when using more contextual features, we did not achieve better performance. Further investigations are required to identify what matters to define the context.

Figure 1 shows the learning dynamics for all bandits across all days in the data sample. To improve interpretation, we include only the best contextual bandit. The Juggler-MAB hybrid demonstrated fast adaptation to changing conditions compared to the original Juggler model. We observed that the MAB component was able to make fine-grained adjustments to the Juggler predictions, resulting in improved performance.

Diving now deeper into the arms selection per bandit, we present Figure 2. The results show a clear and expected preference towards arms lower compensation weights, as they are not aligned with the NDCG reward. However, it is interesting to observe that the best bandit has learned that not only is it ideal to decrease compensation, but also to increase or decrease relevance depending on the context.

We inspect now Juggler-MAB's effect on lodging ranking top-10 average statistics in Table 2. The results are reported as differences to the Juggler baseline, as we cannot exposure the raw data due to its sensitiveness.

| Metric | $\epsilon$-greedy (0.3) | $\epsilon$-greedy (0.1) | RLS |
|---|---|---|---|
| daily price | -0.7278 | -0.8324 | -0.8595 |
| guest rating | 0.0416 | 0.0572 | 0.0604 |
| star rating | 0.0499 | 0.0747 | 0.0796 |
| margin % | -0.0034 | -0.0045 | -0.0048 |
| margin $ | -0.6285 | -0.8222 | -0.8633 |

Table 2: Differences in several metrics for bandits versus the Juggler baseline in top-10 positions.

The results show a clear pattern for all bandits: average daily price decreases and guest and star ratings increase as NDCG improves. On the contrary, margin % and margin $ decreases, which could pose problems to the marketplace objectives and long term health. The expectations, to be validated via AB test, is that the increase in relevance will lead to an improvement in conversion rate which can offset the impact in profit per transaction.

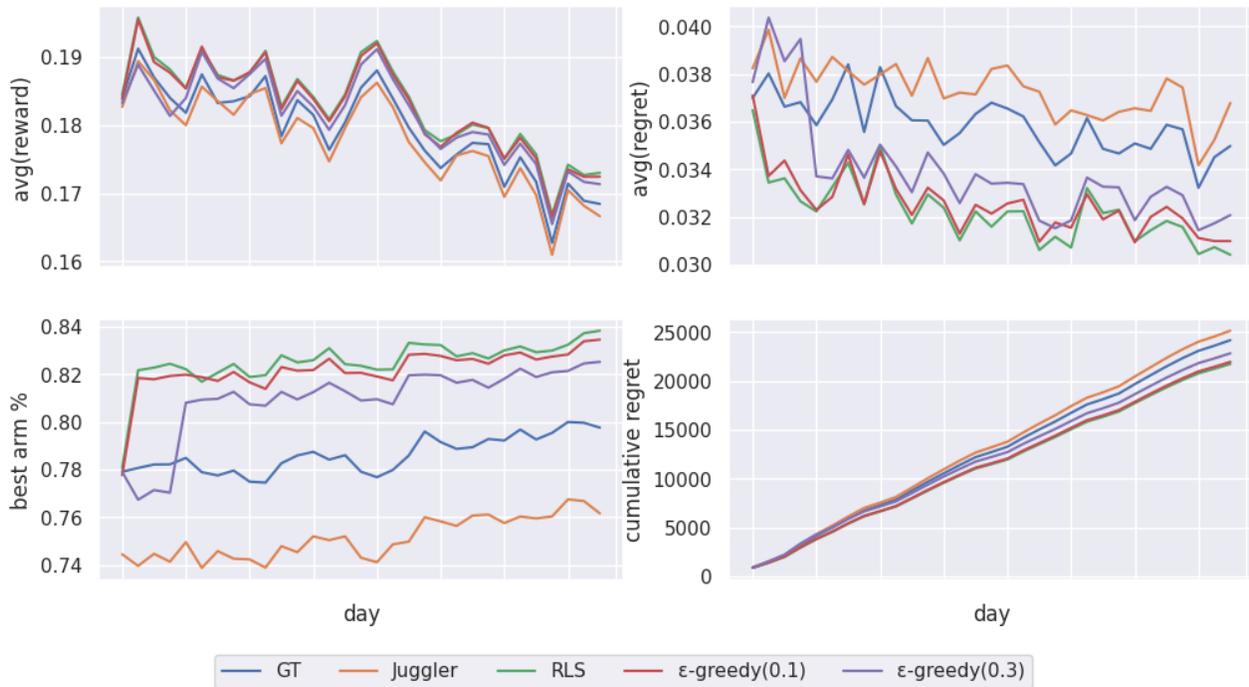

Figure 1: Multiple metrics per bandit over time.

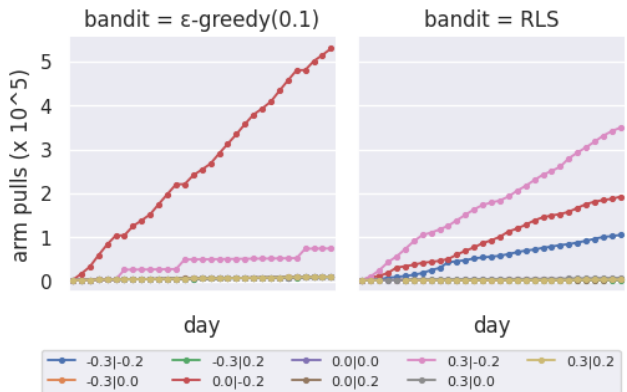

Figure 2: Arm pulls per bandit over time.

Despite the overall positive results, we identified the following limitations and challenges:

(1) The reward function considers only a single dimension of the problem, i.e. relevance. This explains the impact observed to the compensation component. Future work will address this limitation by using multi-objective optimization techniques [13].
(2) Our current simulations use historical interactions with a deterministic logging policy, introducing bias. To address this, we plan to implement off-policy evaluation techniques [18, 19].

## 6 CONCLUSION AND FUTURE WORK

In this paper, we presented a novel hybrid approach combining Meta-Learning with Multi-Armed Bandits for multi-stakeholder recommendations in online travel marketplaces. Our Juggler-MAB system demonstrated significant improvements over existing methods. Key contributions of our work include 1) an integration of meta-learning and contextual bandits for recommendation systems and 2) empirical evidence of the effectiveness of our approach in a large-scale, real-world setting.

Based on our findings and the limitations identified, we propose the following directions for future research:

(1) Online testing: Conduct A/B tests in a production environment to validate the performance of Juggler-MAB under real-world conditions and user behaviors
(2) Dynamic arm space: Explore methods for dynamically adjusting the arm space of the MAB component based on observed performance and changing market conditions.
(3) Integration with other recommendation components: Investigate how Juggler-MAB can be integrated with other components of the recommendation pipeline, such as candidate generation and diversity enhancement.
(4) Fairness considerations: Incorporate explicit fairness constraints or objectives into the MAB formulation to ensure equitable treatment of different provider segments [20]
(5) Long-term value optimization: Extend the approach to consider long-term user value, potentially using reinforcement learning techniques for sequential decision-making.